\documentclass[runningheads]{llncs}

\usepackage[mobile]{eccv}

\usepackage{eccvabbrv}

\usepackage{graphicx}
\usepackage{booktabs}

\usepackage[accsupp]{axessibility}  %

\usepackage{hyperref}

\usepackage{orcidlink}

\usepackage{amssymb}
\usepackage{pifont}
\usepackage{arydshln}

\newcommand{\cmark}{\ding{51}}%
\newcommand{\xmark}{\ding{55}}%

\begin{document}
\emergencystretch 3em

\title{Statewide Visual Geolocalization in the Wild}

\author{Florian Fervers\inst{1,2}\orcidlink{0000-0002-1263-0355} \and
Sebastian Bullinger\inst{1}\orcidlink{0000-0002-1584-5319} \and
Christoph Bodensteiner\inst{1}\orcidlink{0000-0002-5563-3484} \and
Michael Arens\inst{1}\orcidlink{0000-0002-7857-0332} \and
Rainer Stiefelhagen\inst{2}\orcidlink{0000-0001-8046-4945}}

\authorrunning{F.~Fervers et al.}

\institute{Fraunhofer Institute of Optronics, System Technologies and Image Exploitation\\ \email{\{firstname.lastname\}@iosb.fraunhofer.de} \and
Karlsruhe Institute of Technology\\\email{rainer.stiefelhagen@kit.edu}}

\maketitle
\begin{abstract}
	This work presents a method that is able to predict the \mbox{geolocation} of a street-view photo taken in the wild within a \mbox{state-sized} search region by matching against a database of aerial reference imagery. We partition the search region into geographical cells and train a model to map cells and corresponding photos into a joint embedding space that is used to perform retrieval at test time. The model utilizes aerial images for each cell at multiple levels-of-detail to provide sufficient information about the surrounding scene. We propose a novel layout of the search region with consistent cell resolutions that allows scaling to large geographical regions. Experiments demonstrate that the method successfully localizes 60.6\% of all non-panoramic street-view photos uploaded to the crowd-sourcing platform Mapillary in the state of Massachusetts to within 50m of their ground-truth location. Source code is available at \mbox{\url{https://github.com/fferflo/statewide-visual-geolocalization}}.
    \keywords{Visual Geolocalization \and Image Retrieval \and Image Localization}
\end{abstract}
\begin{figure}[th]
	\centering
	\includegraphics[width=\linewidth]{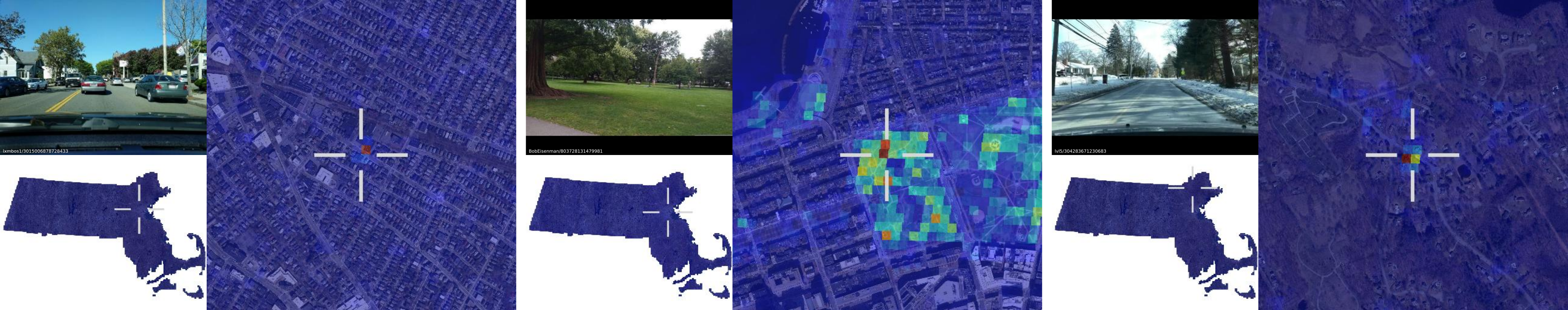}
	\caption{Successful localization of street-view photos in the state of Massachusetts. The search region's color indicates the predicted score for possible camera locations. The crosshair shows the ground-truth location.}
	\label{fig:title}
\end{figure}
\section{Introduction}\label{sec:introduction}
Visual geolocalization (VGL) aims to estimate the geolocation of a photo without relying on additional signals \eg from global navigation satellite systems (GNSS). VGL typically matches the query image against a database of reference images that were captured in the region of interest, and determines the geolocation from visually similar matches.

Two principal lines of research on VGL have emerged that follow different choices for the reference data. Visual place recognition (VPR) utilizes images that were captured from similar viewpoints as the query image \cite{chen2011city,torii201524,arandjelovic2016netvlad,berton2022rethinking,berton2022deep,hausler2021patch,berton2023eigenplaces,ali2023mixvpr}, for instance utilizing platforms such as Google Streetview or Mapillary. \mbox{Close-by} perspectives in the reference data potentially benefit the matching process by providing similar visual cues as the query image. However, VPR is limited to regions where a sufficiently dense cover of street-view images is available. For instance, to perform localization in the city of San Francisco with an area of ${\sim}97\text{km}^2$, recent methods train on a dataset with more than 41M photos from this region and utilize $2.8$M photos as the reference database \cite{berton2022rethinking,berton2023eigenplaces}.

The second line of research - cross-view geolocalization (CVGL) - has opted for aerial imagery as reference data against which the street-view query is matched \cite{deuser2023sample4geo,zhu2022transgeo,zhu2023simple,shi2019spatial,vyas2022gama,zhang2023cross,zhu2021vigor,liu2019lending,vo2016localizing,workman2015wide,rodrigues2022global,shugaev2024arcgeo}. Orthophotos provide a dense and large-scale coverage of outdoor regions beyond street-view databases, and can be captured and kept up-to-date effectively. For instance, aerial imagery for the entire state of Massachusetts was captured in 14 flight days over a span of less than two months \cite{massgis}. Recent works have demonstrated the potential of modern deep learning models to address the drastic changes in viewpoint and scale between the query and reference images. However, they target small, city-sized test regions less than $150\text{km}^2$ in size, and rely on the use of street-view panoramas that provide $360^\circ$ field of view (FOV), on a known compass orientation that eases the matching with \mbox{north-aligned} aerial imagery, or on the availability of street-view imagery from the target region for training \cite{workman2015wide,liu2019lending,zhu2021vigor,zhu2022transgeo,deuser2023sample4geo}.

In this work, we present a comprehensive method that addresses the drawbacks of existing approaches in the field of CVGL and pushes the boundaries of what VGL is capable of. Our contributions are as follows:
\begin{itemize}
	\item We demonstrate the feasibility of localizing street-view photos up to 50m accuracy in the wild (\ie without information about the camera's intrinsics, distortion parameters or orientation) in far larger regions than previously possible (such as the entire state of Massachusetts with ${\sim}23000\text{km}^2$) with limited FOV and without seeing data from the test region during training.
	\item We design a novel model for CVGL that exploits aerial imagery at multiple levels-of-detail (LOD) to address the limited FOV of street-view images and outperforms previous models both in a multiple LOD and single LOD setting.
	\item We present a novel layout of the search region that factors in distortions due to map projections of the earth's surface and allows scaling to large geographical regions.
	\item To evaluate our method, we collect a large and diverse dataset for CVGL consisting of all non-panoramic \mbox{street-view} images uploaded to the \mbox{crowd-sourcing} platform Mapillary across several states in the US and Germany that provide access to statewide aerial imagery.
\end{itemize}

\section{Related Work}
\label{sec:related}
\subsection{Street-view to Street-view Geolocalization}
VPR utilizes \mbox{street-view} images as reference data to localize other street-view query photos. For a clearer distinction, we refer to it as street-view to street-view geolocalization (SVGL). The task is commonly addressed as a retrieval problem where the query's location is determined from the geolocations of the visually most similar reference image(s) retrieved from a database \cite{arandjelovic2016netvlad,hausler2021patch,jin2017learned,xu2023transvlad,berton2022rethinking,berton2023eigenplaces,ali2022gsv,ali2023mixvpr}. A reference image is assumed to be a match if its location is within a fixed distance (\eg 25m \cite{berton2022deep}) to the query's location.

Retrieval-based methods for SVGL typically train a model to map images onto embedding features using a contrastive objective such as the triplet \cite{arandjelovic2016netvlad,jin2017learned} or multi-similarity \cite{ali2022gsv,ali2023mixvpr} loss. The distance between two embeddings reflects the predicted probability that the respective images are captured from the same or a close-by location. At test time, the embeddings are used to perform nearest neighbor search in the database and retrieve matches for each query image.

A different line of research formulates the problem as a classification task: The region of interest is divided into geographical cells each representing a class, and a model is trained to predict the correct class given a street-view image as input \cite{izbicki2020exploiting,weyand2016planet,vo2017revisiting,seo2018cplanet,trivigno2023divide,pramanick2022world,clark2023we}. Most works on classification-based SVGL consider large search regions with a coarse resolution of cells, \eg covering the entire globe with roughly $10^5$ cells \cite{vo2017revisiting,seo2018cplanet}. Recently, Trivigno \etal \cite{trivigno2023divide} demonstrate that classification is also feasible at finer resolutions of $20\text{m}\times20\text{m}$ in a city-sized region. However, they require a sufficient density of training images from the target region and utilize a dataset with an average of more than 100 training photos per cell.

\subsection{Cross-view Geolocalization}
\paragraph{Overview} CVGL utilizes aerial images as reference data to localize street-view images. It is commonly framed as a retrieval problem where a street-view image is matched against a database of georeferenced aerial images to determine its geolocation \cite{lin2015learning,workman2015wide,vo2016localizing,hu2018cvm,liu2019lending,cai2019ground,shi2019spatial,zhu2021vigor,zhu2022transgeo,deuser2023sample4geo,zhang2023cross2,shugaev2024arcgeo}. Unlike in SVGL, the query and reference images are not sampled from the same domain, requiring the model to map images from different domains into a \textit{joint} embedding space. The embedding distance reflects the predicted probability that a query image is located in the region represented by an aerial reference image.
\paragraph{Problem formulation} Most existing benchmarks for CVGL consider a \mbox{one-to-one} matching problem between street-view and aerial images: Datasets are constructed by collecting a set of street-view photos and sampling a single aerial image patch for each photo's location. The resulting reference database thus covers only a sparse set of locations across a test region and prohibits localizing novel photos \cite{workman2015wide,liu2019lending,vo2016localizing} or videos \cite{zhang2023cross,vyas2022gama} at other locations. Furthermore, if every street-view photo is located at the center of the respective aerial image, the model is trained to match reference images only to query photos from a single geo-coordinate \cite{workman2015wide,liu2019lending}. 

To address this limitation, Zhu \etal \cite{zhu2021vigor} introduce the VIGOR dataset which reformulates CVGL as a many-to-one matching task: The search region is covered densely with overlapping aerial image patches, and each image covers a well-defined region of possible camera locations that it is matched with. This facilitates the real-world scenario of localizing novel query photos anywhere inside a predefined target region.

The many-to-one formulation of CVGL resembles the classification-based formulation of SVGL. In both cases, the search region is divided into geographical cells, an embedding is constructed for each cell and compared with the query to determine the matching score. In \mbox{classification-based} SVGL, the embeddings for the test region are learned \textit{during training} as the rows of the last layer's weight matrix and stored as part of the model. In \mbox{retrieval-based} CVGL, embeddings are predicted \textit{after training} using aerial imagery from the test region and stored as the reference database.

All existing cross-view datasets sample aerial images from a Web Mercator projection \cite{epsg3857} that is used for example by Google Maps. The projection causes the scale of images (\ie meters per pixel) to vary based on the image's latitude. This leads to a domain gap when training and testing on different image scales, and to incorrect groundtruth annotations if not accounted for \cite{zhu2021vigor,lentsch2023slicematch}. Our dataset follows the many-to-one formulation of CVGL, but addresses the scale inconsistency of the aerial imagery and search region layout in existing datasets, and introduces the usage of aerial imagery at multiple LOD.

\paragraph{Methods} In recent years, research on CVGL has focused mainly on enhancing a model's ability to predict matching embeddings for query and reference images. A global receptive field in the architecture by means of a multilayer perceptron (MLP) along the spatial dimensions \cite{shi2019spatial,zhu2023simple} or global \mbox{self-attention} \cite{yang2021cross,zhang2023cross2,zhu2022transgeo} enables the model to reason about the spatial layout of the scene. Auxiliary loss functions provide additional supervision during training, \eg by predicting the relative orientation \cite{vo2016localizing,cai2019ground} or metric offset \cite{zhu2021vigor} between \mbox{street-view} and aerial images. Shi \etal \cite{shi2019spatial} apply a polar transformation to the aerial image around the street-view camera's location to bridge the domain gap. However, this is limited to a one-to-one setting where the camera location on the matching aerial image is known in advance, and degrades performance in a \mbox{many-to-one} setting \cite{zhu2022transgeo}.

Recently, Deuser \etal \cite{deuser2023sample4geo} utilize a general-purpose contrastive learning setup for CVGL that is also used in other domains (\eg CLIP \cite{radford2021learning}), with an \mbox{off-the-shelf} vision model \cite{liu2022convnet} and a domain-specific hard example mining strategy. Their method, Sample4Geo, substantially improves the \mbox{state-of-the-art} while only requiring a simple training recipe.

Several related works consider the problem of estimating the 3-DoF camera pose of a street-view image on the matching aerial image \cite{xia2021cross, xia2022visual,zhu2021vigor,shi2022beyond,fervers2023uncertainty,sarlin2023orienternet,sarlin2023snap,shi2023boosting,xia2023convolutional,wang2023view}. They assume that a prior pose is given for instance by \mbox{cross-view} retrieval or GNSS, extract an aerial image at this location and estimate the relative pose of the street-view photo.

\section{Method}
\label{sec:method}
\begin{figure}[t]
	\centering
	\begin{subfigure}[t]{0.49\linewidth}
		\vskip 0pt
		\centering
		\includegraphics[width=0.99\linewidth,trim={0 12mm 0 0},clip]{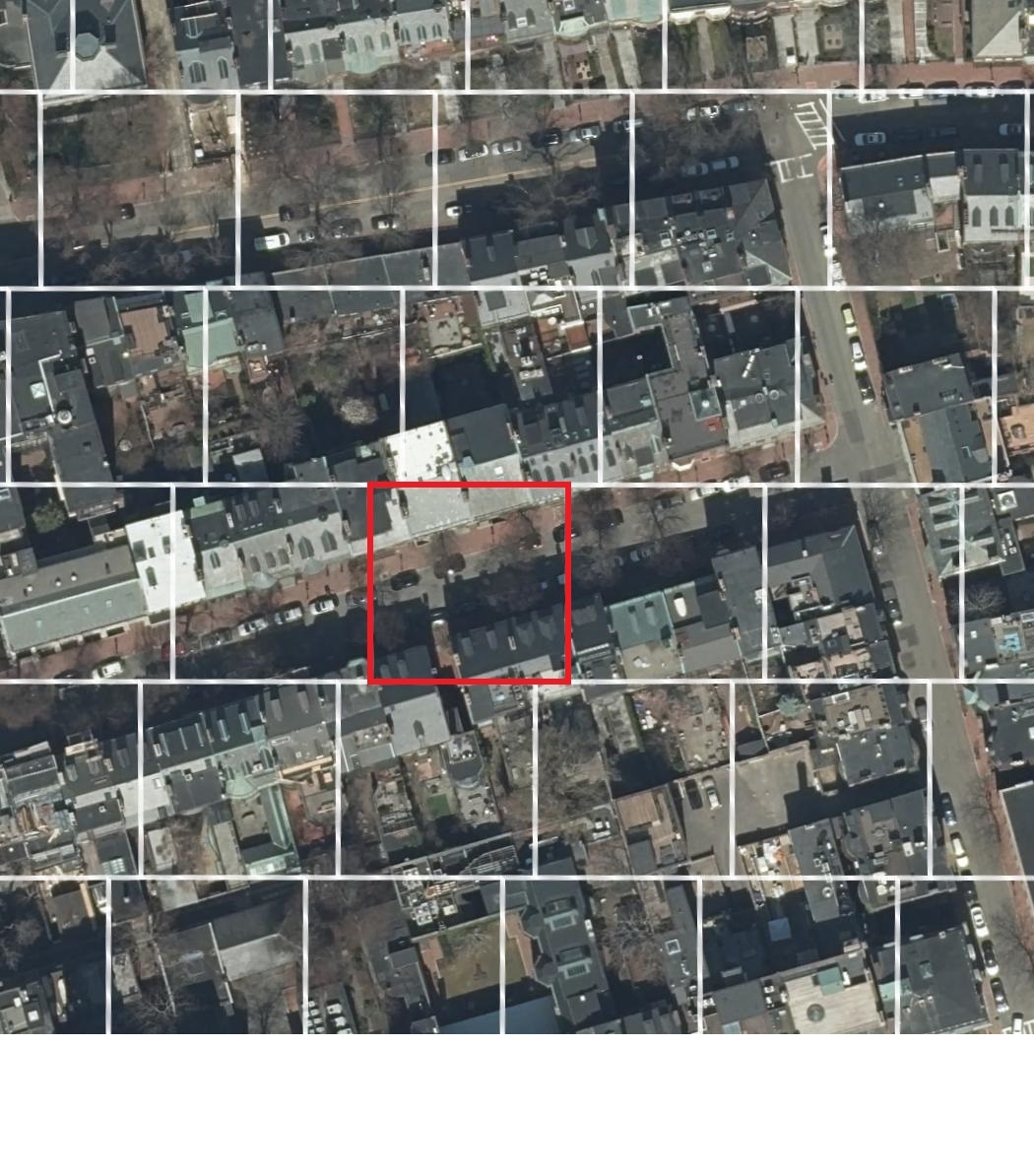}
		\caption{Layout of the search region. Each row of cells corresponds to a longitudinal band around the globe.}
		\label{fig:searchregion-a}
	\end{subfigure}%
	\hfill
	\begin{subfigure}[t]{0.49\linewidth}
		\vskip 0pt
		\centering
		\includegraphics[width=0.99\linewidth,trim={0 5.5mm 0 0},clip]{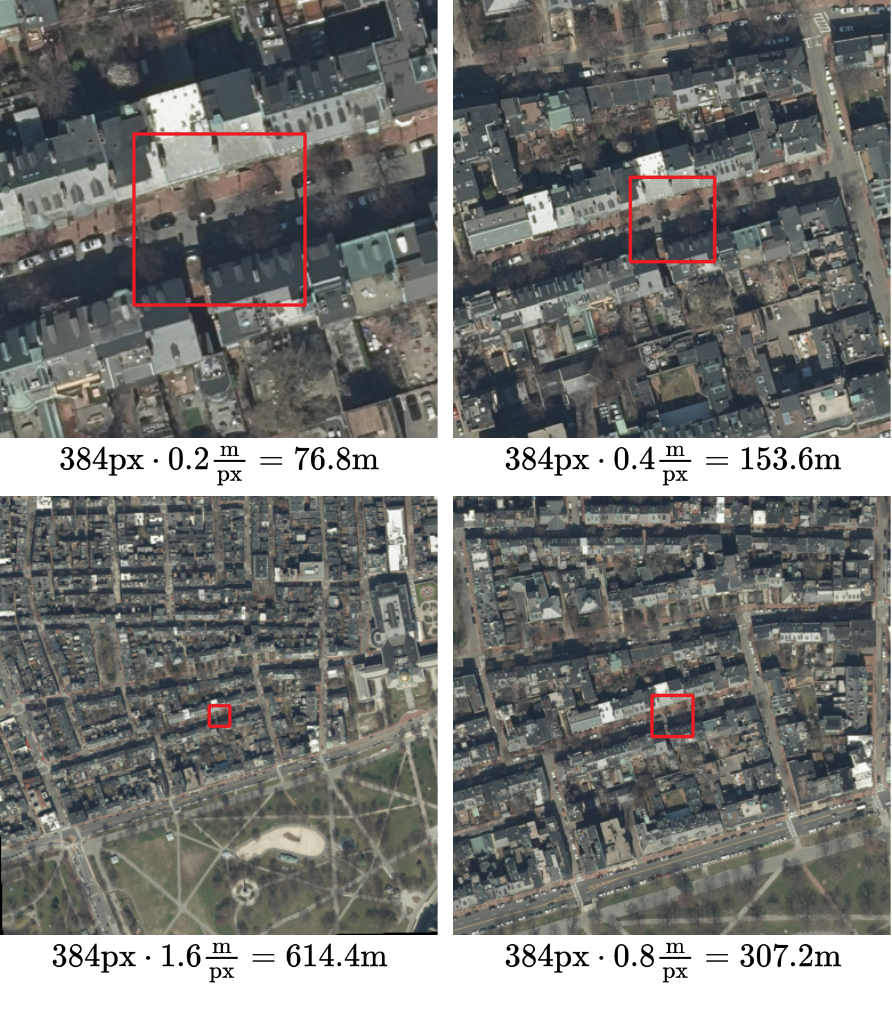}
		\caption{Aerial images passed to the model for the red cell with $384\times384$ pixels and varying metric resolutions.}
		\label{fig:searchregion-b}
	\end{subfigure}
	\caption{Example of the search region layout. Each box in (a) represents a search region cell and is assigned an embedding vector. The embedding is predicted using multiple resolutions of aerial imagery centered on the cell. This provides higher detail for parts of the scene that are close to the street-view camera, and less detail for parts that are further away and appear smaller in the photo.}
\end{figure}
Our method estimates the geolocation of a street-view photo by matching against a database of aerial reference imagery. We consider a state-sized search region (\eg Massachusetts with ${\sim}23000\text{km}^2$) and divide it into a set of geographical cells (\cf \cref{fig:searchregion-a}). Given a street-view photo, our method aims to retrieve the matching cell and thereby provide an estimate of the photo's geolocation. We consider the localization of all photos \textit{in the wild}, \ie with varying and unknown camera intrinsic and distortion parameters, and no information about the camera orientation such as a compass angle or the direction of gravity.

We train a model on a large and diverse dataset to map street-view photos and geographical cells into a joint embedding space using a constrastive learning objective. The resulting embedding distance reflects the probability that a street-view camera is located within a given search region cell. To predict the embedding of a geographical cell, we utilize multiple overlapping aerial images with varying coverage and LOD of the scene around the cell (\cf \cref{fig:searchregion-b}) that provide sufficient overlap with street-view photos taken from within the cell.

To localize photos in a test region, we predict and store embeddings for all cells as our reference database. Given a \mbox{street-view} query image, we use its embedding to perform a nearest neighbor search in the database and retrieve matching cells as well as their matching scores.

\subsection{Search Region Layout}\label{sec:search-region}
We choose a division of the search region into cells such that every cell has roughly the same shape and total area. This enables the model to specialize on a single type of cell regardless of its geographical location, and assigns the same prior probability to all parts of the search region.

A naive division of the region into a regular two-dimensional grid under these constraints is not possible due to the curvature of the earth's surface. VIGOR \cite{zhu2021vigor} instead applies a regular grid to a Web Mercator projection \cite{epsg3857} to define the search region cells. The projection provides a flattened view of the sphere's surface that inflates areas at large distances to the equator. Since cells in this layout are defined to have a constant size in projected coordinates, their metric size in sphere coordinates therefore decreases with their distance to the equator. This results in (1)~a higher prior probability for regions at larger latitudes due to the higher density of cells and (2)~a scale inconsistency of the aerial images that are chosen with a size proportional to the cell size.

To address the scale inconsistency, we choose the search region layout as follows. We assume a spherical model of the earth with radius $r$ and discretize the latitudinal axis with a step size of $l$ meters (\ie $\frac{l}{r}$ radians). This results in longitudinal bands around the globe, where the $i$-th band north or south of the equator is centered on a latitude of $\phi_i = i\frac{l}{r}$ radians.

At each latitude $\phi_i$, we discretize the longitudinal axis with a step size of $l$ meters. Since the longitudinal circumference is reduced by a factor of $\cos{\phi_i}$ \wrt the equator, there are fewer total steps and a larger angle of $\frac{l}{r_i}$ radians per step with $r_i = r\cos{\phi_i}$. The $j$-th step east or west of the prime meridian is located at a longitude of $\lambda_{ij} = j\frac{l}{r_i}$ radians.

We define cells as north-aligned squares of size $l \times l$ centered on the coordinates $(\phi_i, \lambda_{ij})$. \Cref{fig:searchregion-a} shows an example of the resulting search region layout. The longitudinal bands appear shifted \wrt each other due to the different longitudinal circumference and number of cells. At the scale of individual cells the distortion resulting from the globe's curvature is negligible and the scene appears locally flat. We provide an estimate of the approximation error in the supplementary material.
\subsection{Choice of Aerial Images per Cell}
\begin{figure}[t]
	\definecolor{red}{rgb}{1.0, 0.0, 0.0}
	\definecolor{green}{rgb}{0.0, 0.79, 0.0}
	\definecolor{blue}{rgb}{0.0, 0.0, 0.79}
	\centering
	\begin{subfigure}{0.32\linewidth}
		\centering
		\includegraphics[width=0.9\linewidth]{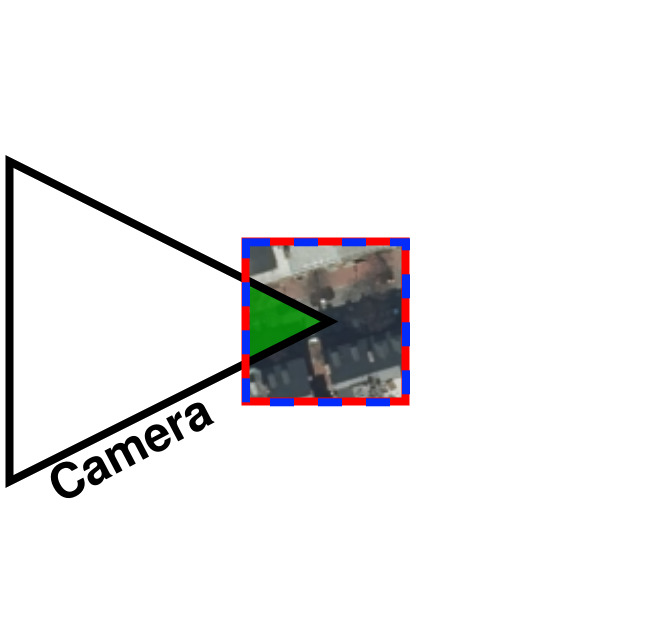}
		\caption{\textit{Naive}: A single aerial image matching the cell size.}
		\label{fig:im-per-cell-a}
	\end{subfigure}
	\hfill
	\begin{subfigure}{0.32\linewidth}
		\centering
		\includegraphics[width=0.9\linewidth]{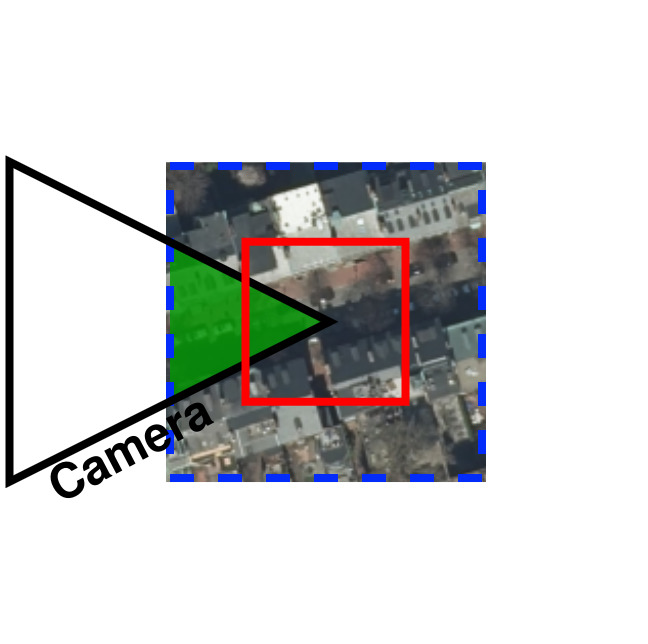}
		\caption{\textit{Ours}: Example with $n=1$ aerial image.}
		\label{fig:im-per-cell-b}
	\end{subfigure}
	\hfill
	\begin{subfigure}{0.32\linewidth}
		\centering
		\includegraphics[width=0.9\linewidth]{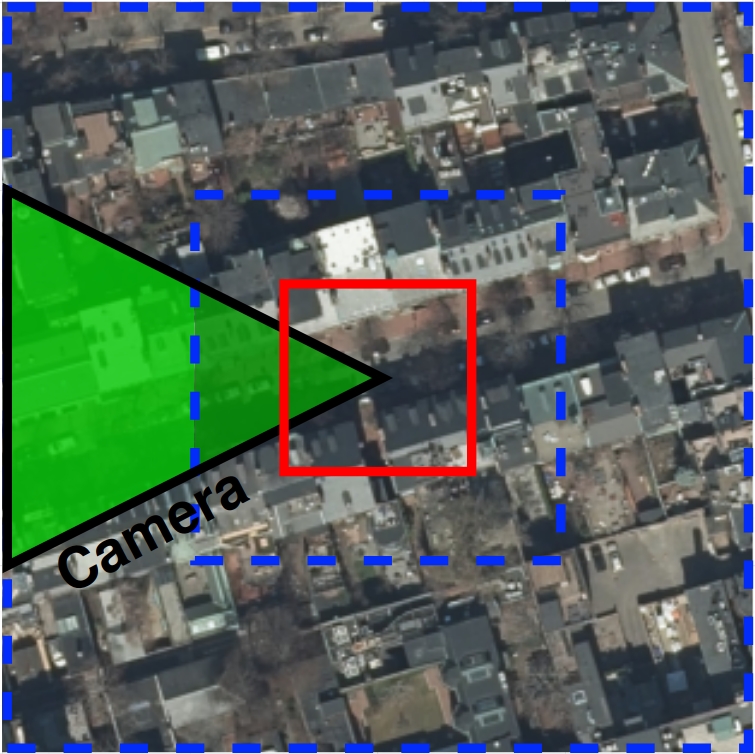}
		\caption{\textit{Ours}: Example with $n=2$ aerial images.}
		\label{fig:im-per-cell-c}
	\end{subfigure}
	\caption{Different choices of aerial images per search region cell. The \textcolor{red}{solid square} delinates the search region cell. We position an example street-view camera with limited FOV at the center of the cell. A \textcolor{blue}{dashed square} represents a single aerial image. The \textcolor{green}{highlighted triangular region} indicates the parts of the aerial image(s) that overlap with the camera frustrum.}
	\label{fig:im-per-cell}
\end{figure}
Given a division of the search region into cells, we extract aerial images for all cells to predict their embedding representations. Ideally, the aerial imagery used for a cell depicts large parts of the surrounding scene that are also visible from camera poses within the cell. This allows the model to exploit sufficient visual cues in the scene and predict matching embeddings for the street-view photo and search region cell.

Naively choosing a single aerial image matching the size of the cell results in little overlap with the camera, especially if it is located near the cell boundary and faces away from the cell (\cf \cref{fig:im-per-cell-a}). The VIGOR benchmark \cite{zhu2021vigor} includes a single aerial image at twice the sidelength of the corresponding cell. However, the use of panoramas as street-view images still ensures that part of the viewing frustrum always faces the cell center and provides a large overlap with the aerial image. To improve the matching of images with limited FOV, we increase the size of the area covered by the aerial reference imagery of a single cell as follows.

Parts of the scene that are further away from the camera are always depicted with less detail in the street-view photo, but have a larger footprint in the aerial imagery. Given this observation, we choose a set of $n$ aerial images per cell that have the same size in pixels, but provide a higher LOD for regions close to the cell, and a lower LOD for larger regions further from the cell (\cf \cref{fig:im-per-cell-c}). The metric sidelength is $d_0$ for the first image, and doubled for every subsequent image:
\begin{equation}
	d_i = 2^i d_0 \text{ \hspace{0.3cm}for } i \in [0..n-1]
\end{equation}
The value $d_0$ is a tunable hyperparameter. \Cref{fig:searchregion-b} shows an example with $n = 4$ and $d_0 = 384\text{px} \cdot 0.2 \frac{\text{m}}{\text{px}} = 76.8\text{m}$.

Using our multi-scale reference imagery requires significantly less memory and computational resources than a single reference image with high LOD and coverage. We pass all images jointly to the model to predict a single embedding vector for the cell.

\subsection{Model}

\begin{figure}[t]
	\centering
	\includegraphics[width=0.99\linewidth]{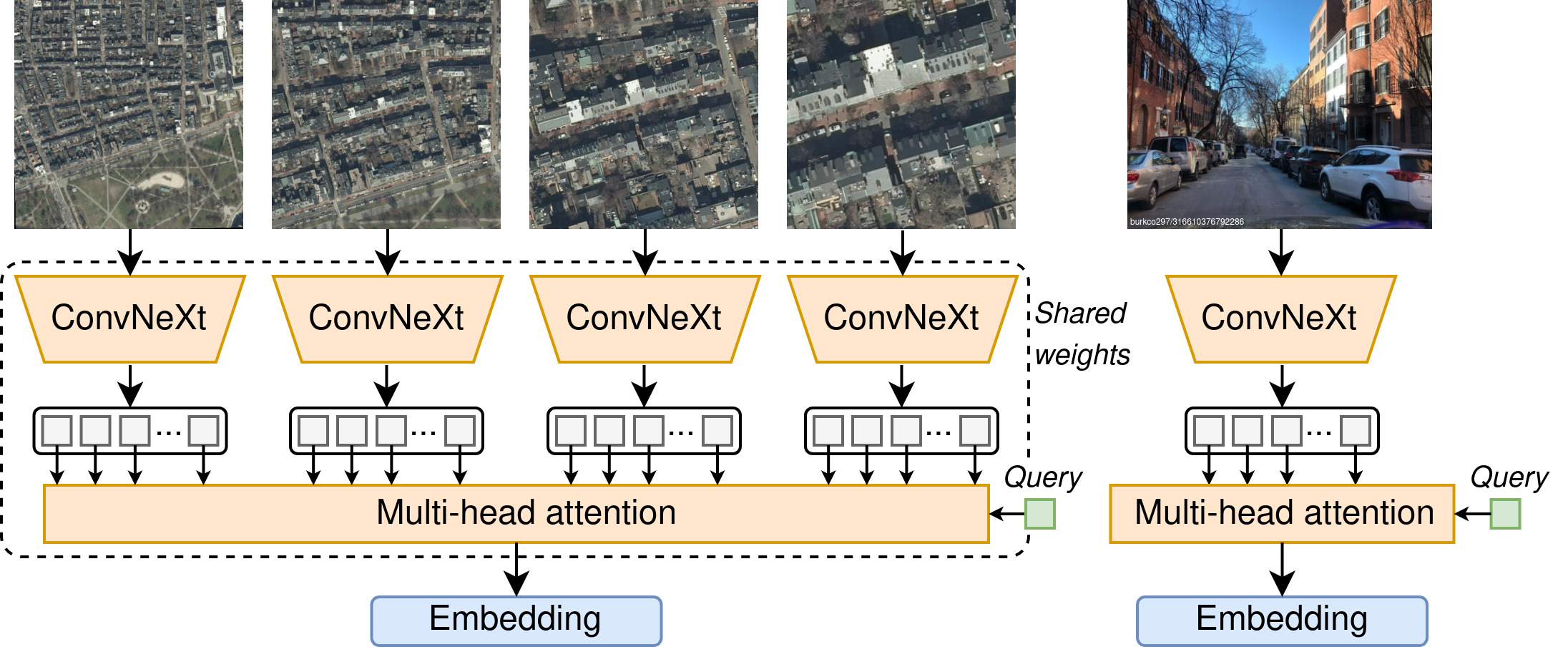}
	\caption{Overview of the architecture. Our model uses the ConvNeXt backbone \cite{liu2022convnet} to encode images, and a multi-head attention block to pool the encoded tokens into a single embedding representation \cite{vaswani2017attention,lee2019set}. We use shared weights for the model applied to the aerial images.}
	\label{fig:model}
\end{figure}

We design a model to map street-view images and search region cells into a joint embedding space that is used to retrieve matching cells at test time (\cf \cref{fig:model}). We employ the ConvNeXt backbone \cite{liu2022convnet} to encode the input images and an attention-based pooling layer \cite{vaswani2017attention,lee2019set} to aggregate the encoded features resulting in an embedding vector for the street-view photo and the search region cell.

We use separate encoders for the street-view and aerial domains with the same architecture, but different weights. In the aerial domain, we apply the same encoder in batched mode to all images with different LOD which reduces the memory and computational requirements compared to using separate models for each LOD. The encoder predicts a feature map for each image which we flatten along the spatial dimensions resulting in a list of tokens.

The pooling layer aggregates information from all encoded tokens into an embedding vector. In the street-view domain the tokens originate from the single input image, while in the aerial domain the tokens from multiple reference images are concatenated first. We use a multi-head attention (MHA) block \cite{vaswani2017attention} with $h$ heads and a single learnable query token that attends to the encoded image tokens and results in a single output vector. The attention block thus performs token pooling both per image and across images. The output vector is \mbox{L2-normalized} to yield the final embedding.

\subsection{Training Setup}
\label{sec:training-setup}

We train the model in a contrastive setting similar to Sample4Geo \cite{deuser2023sample4geo} and CLIP \cite{radford2021learning}, but use the decoupled contrastive loss (DCL) \cite{yeh2022decoupled} to alleviate the use of a smaller batch size. For each training batch, we sample $b$ \mbox{street-view} images from the dataset and $b$ matching search region cells. To reduce the oversampling of regions in the dataset with many densely captured street-view images, we partition all photos into cells of size $5\text{m} \times 5\text{m}$ following the layout described in \cref{sec:search-region}. For each batch, we sample $b$ such cells and include one photo per cell.

For each street-view photo in the batch, the matching cell is constructed with a random orientation $\theta \in [0, 2\pi)$ and offset $t \in [-t_\text{max}, t_\text{max}]^2$ to the camera location with \mbox{$t_\text{max} = 0.5l - l_\Delta$}. We choose $l_\Delta = 5\text{m}$ to partially compensate for errors in the camera's GNSS position. The random orientation during training serves as a method for data augmentation, \eg to generalize over the direction of shadows in the scene or an oblique camera angle.

We compute the cosine similarity between the embeddings of all street-view images and search region cells in the batch resulting in $b^2$ matching scores. We apply a symmetric DCL to maximize the similarity of the $b$ positive pairs and minimize the similarity of the $b^2 - b$ negative pairs. To exclude false negatives, we mask out pairings where the \mbox{street-view} image has a distance of less than $100$m to the negative search region cell.%

\subsection{Hard Example Mining}

When using a random sampling strategy as described above, the model tends to classify most samples per batch correctly after some time during the training and receives little more supervision from the batch. While some works on contrastive learning afford large batch sizes to address this problem \cite{radford2021learning}, we use a hard example mining (HEM) strategy with a smaller batch size instead.

Existing works in the field of VGL store the embeddings predicted for the training split in memory to mine hard examples for each batch. The embeddings are either updated using each training batch \cite{zhu2021vigor,zhu2022transgeo}, or by periodically running inference on the training split \cite{deuser2023sample4geo,arandjelovic2016netvlad} or a subset that is used as the mining pool \cite{warburg2020mapillary}. Unlike existing works, we consider a semi-infinite data regime where the dataset is large enough for every street-view image to be seen only once, and the training consequently encompasses only a single epoch. In this setup, using past training batches or an inference on the entire dataset to predict embeddings for the mining pool similar to previous works is not feasible.

To perform HEM at some point during training, we scan the next $s$ samples (\ie positive query-reference pairs) and predict their embeddings using the current model state. We apply a simple clustering method over the embeddings to partition the mining pool into $\frac{s}{b}$ batches. Each batch consists of a cluster of samples that tend to be difficult for the model to discriminate due to their proximity in the embedding space. To construct a cluster, we select a random sample from the pool and iteratively add $b - 1$ samples whose reference embedding is closest to the cluster's centroid, \ie the mean of the query embeddings. This process is repeated until all samples in the pool are assigned to a cluster. The batches are then forwarded to the original training loop.

We start with $s=b$ (\ie random sampling) at the beginning of the training, and increase $s$ by a factor of 2 after each scanned mining pool (and at least $\frac{5000}{b}$ iterations) up to a maximum of $s_\textit{max}$. Small scan sizes are sufficient to mine hard negatives early during training since the model weights still change rapidly and the recall per batch is low. As the model's performance improves and the weights become more stable, we increase $s$ to find increasingly harder samples in the mining pool.

Our method for HEM resembles a curriculum learning strategy \cite{bengio2009curriculum} where the training progresses from easy to more complex examples. Similar to existing works, it requires an additional forward pass per sample and loading every sample from disk twice when $s$ is large.

\begin{table}[t]
	\newcommand{\spacing}{\hspace{3mm}}
	\small
	\centering
	\caption{Comparison of visual geolocalization datasets. We show only SVGL datasets that include a dense coverage of the test region \cite{berton2022rethinking}. The total geographical coverage is approximated by counting the number of $100\text{m}\times100\text{m}$ cells (\cf \cref{sec:search-region}) that contain at least one street-view photo. We include the size of the test region if it is covered densely with reference imagery, and \xmark\hspace{0.5mm} to indicate that aerial images are provided sparsely for a set of street-view photos. The size of test regions in SVGL datasets is approximated as the number of cells times the cell size. \label{tab:datasets}}
	\begin{tabular}{l|r|r|c|r|c}
		& \# Photos & \# Cells & Videos & Test region & Consistent scale\\
		\hline
		\textbf{Cross-view} &&&&& \\
		\spacing \textit{Ours} & 72.7M & 1.5M & \cmark & 22922.3km$^2$ & \cmark \\
		\spacing CVUSA \cite{workman2015wide} & 44k & 43k & \xmark & \xmark & \xmark \\
		\spacing CVACT \cite{liu2019lending} & 138k & 12k & \xmark & \xmark & \xmark \\
		\spacing Vo \& Hays \cite{vo2016localizing} & 450k & - & \xmark & \xmark & \xmark \\
		\spacing VIGOR \cite{zhu2021vigor} & 105k & 12k & \xmark & 114.9km$^2$ & \xmark \\
		\spacing SeqGeo \cite{zhang2023cross} & 119k & 10k & \cmark & \xmark & \xmark \\
		\spacing Gama \cite{vyas2022gama} & 53.9M & 65k & \cmark & \xmark & \xmark \\
		\hline
		\textbf{Street-view to street-view} &&&&& \\
		\spacing St Lucia \cite{milford2008mapping} & 33k & $<$1k & \cmark & ${<}4.8\text{km}^2$ & - \\
		\spacing NCLT \cite{carlevaris2016university} & 3.8M & $<$1k & \cmark & 0.4km$^2$ & - \\
		\spacing Pittsburgh 250k \cite{arandjelovic2016netvlad} & 274k & $<$1k & \xmark & 0.9km$^2$ & - \\
		\spacing TokyoTM/247 \cite{torii201524} & 288k & $<$1k & \xmark & 4.2km$^2$ & - \\
		\spacing SF Landmarks \cite{chen2011city} & 1.1M & $<$1k & \xmark & 1.3km$^2$ & - \\
		\spacing SF-XL \cite{berton2022rethinking} & 41.2M & 11k & \xmark & 97.0km$^2$ & - \\
	\end{tabular}
\end{table}

\section{Evaluation}
\subsection{Data}
\label{sec:dataset}

Most existing datasets in the field of CVGL \cite{workman2015wide,liu2019lending,vo2016localizing,zhang2023cross,vyas2022gama} represent a \mbox{one-to-one} matching task without a dense coverage of aerial imagery over a test region. VIGOR \cite{zhu2021vigor} includes such coverage and follows a many-to-one formulation of CVGL. However, similar to other datasets it provides a set of preselected aerial image patches with inconsistent scale that limits further research into the search region layout and use of multi-scale reference imagery.

To evaluate our method, we therefore collect a new dataset as follows. We select four states in Germany and three states in the United States that offer statewide access to high-resolution orthophotos. We provide an interface to sample aerial images patches from these sources on-the-fly with the requested size, orientation and scale.

We collect street-view images for these states from the crowd-sourcing platform Mapillary. We download all non-panoramic images in the respective regions that provide meta-data such as their geolocation. The resulting data cover a wide range of geographic locations, different camera models and orientations, weather, daylight and seasonal variations, and are captured under real-world conditions from different sources, including pedestrians, cyclists and motor-vehicles \mbox{(\cf \cref{fig:eval})}. We do not use the camera's intrinsics or distortion parameters, and do not rely on preprocessing steps such as structure-from-motion \cite{sarlin2023orienternet} or \mbox{pseudo-labels \cite{fervers2023uncertainty}} to provide an accurate geolocation and camera orientation. We resize and pad all downloaded photos to a resolution of $640 \times 480$. We observe that the geolocations are often inaccurate and can deviate by tens of meters.

We use data from the state of Massachusetts as test split, and the remaining six states as training split. The test split covers an area of ${\sim}23000\text{km}^2$ and contains ${\sim}11$M street-view images as queries. We use a smaller test split for the ablation studies covering ${\sim}100\text{km}^2$ with $100$k street-view images, similar in size to VIGOR \cite{zhu2021vigor}. %

\Cref{tab:datasets} provides a comparison with existing datasets in the fields of SVGL and CVGL. Our dataset covers a dense test region that is more than $100\times$ larger, and an overall geographical coverage that is more than $25\times$ larger than existing datasets. We provide download scripts for the street-view and aerial data including a list of Mapillary image IDs used in this work. The data is hosted by the Mapillary service and the state's orthophoto providers and is subject to changes on these platforms, \eg due to users deleting their street-view photos.

\if{false}
We summarize the main differences as follows:
\begin{itemize}
	\item Our dataset covers a dense test region that is more than $100\times$ larger, and an overall geographical coverage that is more than $25\times$ larger than existing datasets.
	\item Our dataset includes ${\sim}$73M street-view images spread over several geographic regions. While SF-XL \cite{berton2022rethinking} and Gama \cite{vyas2022gama} contain a similar number of street-view photos, they cover much smaller regions such as the city of San Francisco \cite{berton2022rethinking} or trajectories where many consecutive frames are located on a single aerial image patch \cite{vyas2022gama}.
	\item Our dataset allows sampling aerial images at any size, orientation and scale, allowing for research into the search region layout and use of multi-scale reference imagery.
\end{itemize}
\fi
\subsection{Metric}
Retrieval methods for CVGL are typically evaluated using the recall@$N$ (R@$N$) metric that measures the proportion of queries where the true search region cell is one of the top $N$ retrieved cells \cite{zhu2022transgeo,deuser2023sample4geo,zhu2023simple,shi2019spatial}. However, this metric does not consider (1)~the size and number of search region cells, (2)~the discretization of the search region that would penalize misassignment of query images close to a cell boundary and (3)~inaccurate geolocations in the \mbox{crowd-sourced} dataset that lead to incorrect ground-truth cells.

Since we are interested in providing a rough location estimate for an image rather than the exact cell assignment, we instead consider cells to be a correct match if their center is within a distance of 50m to the camera location (R@$N{<}50$m). Similar metrics based on a localization radius are also used in the context of SVGL \cite{berton2022rethinking,trivigno2023divide}. The choice of 50m allows \eg for a subsequent approximation of the exact 3-DoF camera pose using recent metric CVGL approaches \cite{fervers2023uncertainty}.

\subsection{Implementation}\label{sec:implementation}
We use the ConvNeXt-B encoder \cite{liu2022convnet} pretrained on ImageNet \cite{deng2009imagenet} and apply the MHA layer with $64$ heads. Unless stated otherwise, we use an embedding dimension of $1024$, search region cells with size $30\text{m} \times 30\text{m}$ and four aerial images per cell with $384 \times 384$ pixels and $d_0 = 384\text{px} \cdot 0.2 \frac{\text{m}}{\text{px}} = 76.8\text{m}$. We train for 200k iterations with a batchsize of 30, a learning rate of $1 \times 10^{-4}$ with 1k iterations of linear warmup and the cosine decay schedule. We use label smoothing of $\epsilon = 0.1$ and a temperature of $\tau = \frac{1}{36} \approx 0.03$ in the loss function. The mining pool size is capped at $s_{\text{max}} = 2^{14}$.

For the ablation studies, we choose a training setup with reduced computational cost. We use the ConvNeXt-S encoder, downsize street-view images to $240\times320$ pixels and aerial images to $256\times256$ pixels and cap the mining pool to $s_{\text{max}} = 2^{12}$ samples.

We use Hierarchical Navigable Small Worlds \cite{malkov2018efficient} implemented in the Faiss library \cite{douze2024faiss} to perform approximate nearest neighbor search on the test split, and a simple brute-force approach for the ablation studies.

\subsection{Results}
\begin{figure}[t]
	\centering
	\includegraphics[width=\linewidth]{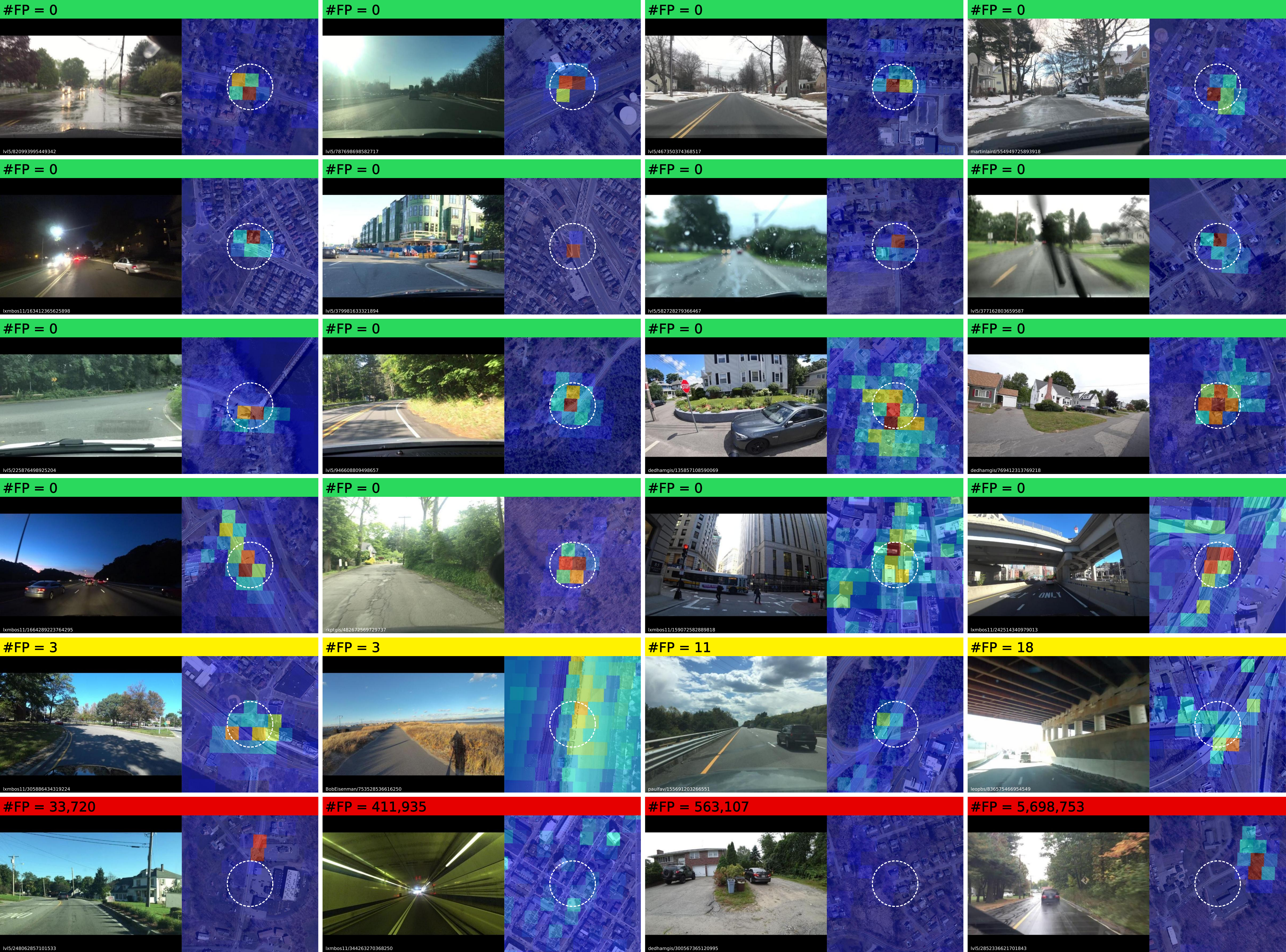}
	\caption{Localization of street-view photos in the state of Massachusetts. Overall, 60.6$\%$ of images are localized correctly. False positives (FP) are cells that are scored higher by the model than all cells within 50m of the ground-truth location. The search region's color indicates the predicted score for possible camera locations. The white circle delinates the 50m radius around the ground-truth position.}
	\label{fig:eval}
\end{figure}
Our method achieves a recall of 60.6\% on the Massachusetts test split, \ie non-panoramic street-view images uploaded to Mapillary in a region of ${\sim}23000\text{km}^2$ (\cf \cref{tab:results}). \Cref{fig:eval} shows example localizations of images and the predicted scores for cells in the search region. The method successfully localizes images even under challenging circumstances such as rain and snow, varying illuminations and perspectives, occlusion, noise, and in regions with few man-made structures.

We reimplement SAFA \cite{shi2019spatial}, TransGeo \cite{zhu2022transgeo} and Sample4Geo \cite{deuser2023sample4geo} as baselines and evaluate them with a Mercator-based cell layout and a single aerial image per cell as in VIGOR \cite{zhu2021vigor}. A full description on how the methods are adapted to our problem setup is provided in the supplementary material. The best existing method, Sample4Geo, achieves a recall of 12.8\% on the test split.

\begin{figure}[t]
	\begin{minipage}[t]{.64\textwidth}
		\includegraphics[width=0.49\linewidth]{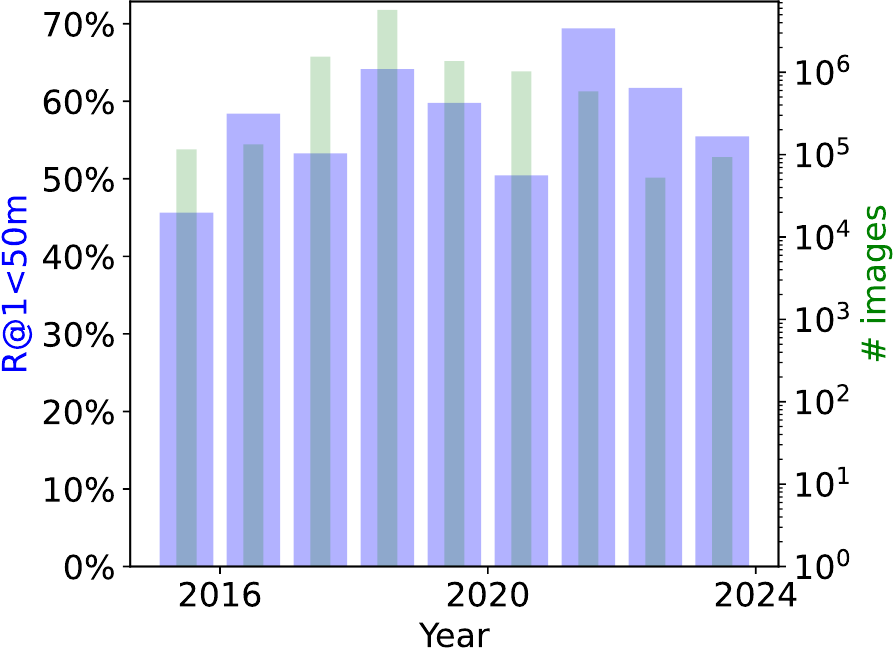}%
		\hfill%
		\includegraphics[width=0.49\linewidth]{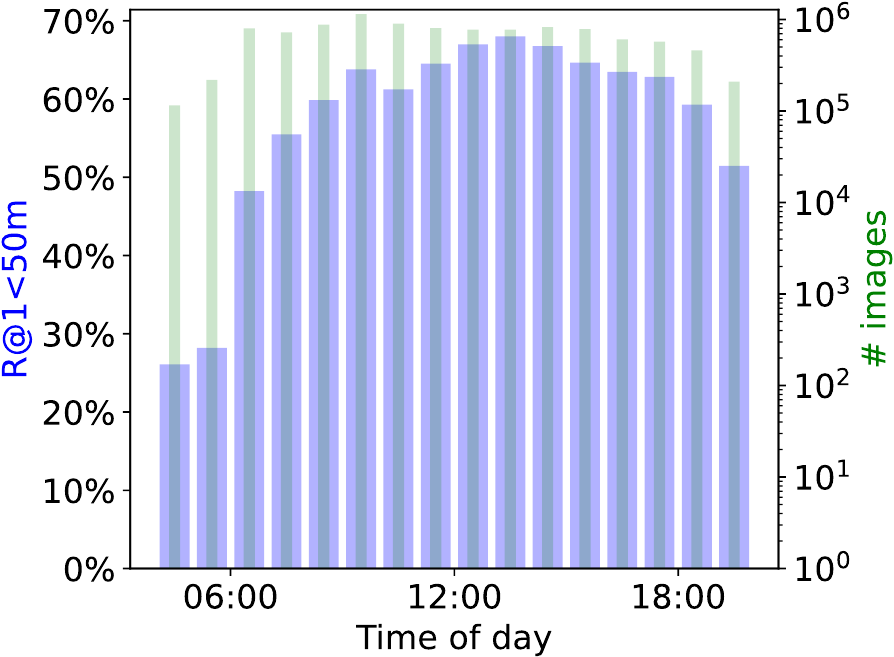}
		\captionof{figure}{Recall for images in the Massachusetts test split captured in different years and during different times of the day. Aerial reference imagery was captured in 2021 \cite{massgis}. We only show histogram buckets with at least 50k images.}\label{fig:time}
	\end{minipage}%
	\hfill%
	\begin{minipage}[t]{.32\textwidth}
		\includegraphics[width=0.98\linewidth]{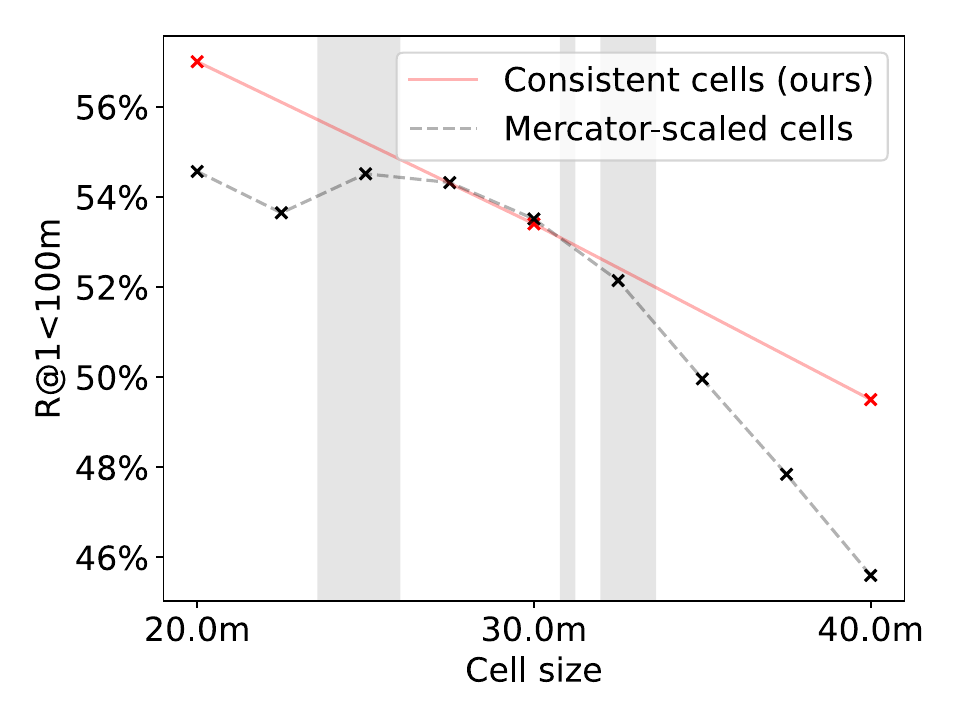}
		\captionof{figure}{Ablation study on different cell sizes. The highlighted regions indicate the size of training cells in a Mercator-based layout.}\label{fig:searchregion}
	\end{minipage}
\end{figure}

To inspect factors that affect the localization performance, we evaluate the recall \wrt time of day and year at which the \mbox{street-view} images were captured. \Cref{fig:time} shows that the recall is highest for photos from 2021 - \ie when the aerial reference data was recorded in Massachusetts \cite{massgis} - and drops when increasing the temporal gap between the capture of query and reference data. The method still correctly localizes 45.7\% of all queries in 2015, highlighting the ability of the model to generalize across temporal changes in the scene. The model further achieves better recall on images captured during the day rather than at night time, potentially due to lower illumination and less training data for this timespan. The method correctly localizes 62.2\% of queries between 06:00 and 18:00, and 46.6\% of queries between 18:00 and 06:00.
\begin{table}[t]
	\fontsize{8pt}{8pt}\selectfont
	\setlength{\tabcolsep}{2pt}
	\centering
	\caption{Recall on the Massachusetts test split in percent. We reimplement three existing methods and evaluate them with a Mercator-based cell layout and a single aerial image per cell as in VIGOR \cite{zhu2021vigor}. More detail on how the baselines are adapted to our problem setup is provided in the supplementary material.}
	\begin{tabular}[t]{l|r|r|r}
		\hline
		\rule{0pt}{2.3ex} & R@$1{<}50$m & R@$10{<}50$m & R@$100{<}50$m \\
		\hline
		\rule{0pt}{2.3ex}SAFA \cite{shi2019spatial} & 0.5 & 2.7 & 10.7 \\
		\rule{0pt}{2.3ex}TransGeo \cite{zhu2022transgeo} & 2.8 & 10.8 & 28.3 \\
		\rule{0pt}{2.3ex}Sample4Geo \cite{deuser2023sample4geo} & 12.8 & 32.2 & 55.8 \\
		\rule{0pt}{2.0ex}\textbf{Ours} & \textbf{60.6} & \textbf{75.5} & \textbf{84.1} \\
		\hline
	\end{tabular}\label{tab:results}
\end{table}

\subsection{Ablation Studies}

For the ablation studies, we train the model in a setup with reduced computational cost (\cf \cref{sec:implementation}) and evaluate on a smaller test split that is similar in size to VIGOR (\cf \cref{sec:dataset}).

\Cref{tab:aerial-lod} shows the localization performance for different choices of aerial images per cell. Using our multi-scale reference imagery yields a recall of 48.3\% compared to 33.4\% when using a single image at twice the sidelength of the cell as in VIGOR \cite{zhu2021vigor}. For fair comparison, we evaluate the single LOD setting with a resolution of $512 \times 512$ pixels, and the multiple LOD setting with four images of $256 \times 256$ pixels, resulting in the same number of input pixels and similar computational cost.
\begin{table}[t]
	\fontsize{6pt}{6pt}\selectfont
	\setlength{\tabcolsep}{2pt}
	\def\arraystretch{0.8}
	\parbox[t][][t]{.35\linewidth}{%
		\centering
		\caption{Comparison of different choices of aerial images per cell.}
		\begin{tabular}[t]{l|r|r}
			\hline
			\rule{0pt}{2.2ex}Meters per pixel & Pixels & R@$1{<}50$m \\
			\hline
			\rule{0pt}{2.6ex}0.12 \cite{zhu2021vigor} & $512^2$ & 33.4\% \\
			0.2 & $512^2$ & 38.9\% \\
			0.4 & $512^2$ & 40.6\% \\
			1.6 & $512^2$ & 28.6\% \\
			0.2, 0.4, 0.8, 1.6 & $4\cdot256^2$ & 48.3\% \\
			0.3, 0.6, 1.2, 2.4 & $4\cdot256^2$ & 46.8\% \\
			0.4, 0.8, 1.6, 3.2 & $4\cdot256^2$ & 46.1\% \\
			\hline
		\end{tabular}
		\label{tab:aerial-lod}
	}%
	\hfill%
	\parbox[t][][t]{.35\linewidth}{%
		\centering
		\caption{Comparison of different pooling layers (designed for single LOD and multiple LOD) with the same encoder.}
		\vspace{3.3mm} %
		\begin{tabular}[t]{l|r|r}
			\hline
			\rule{0pt}{2.2ex}& \multicolumn{2}{c}{R@$1{<}50$m} \\
			Pooling layer & Multi LOD & Single LOD \\
			\hline
			\rule{0pt}{2.2ex}MHA (Ours) & 48.3\% & 38.9\% \\
			Mean \cite{deuser2023sample4geo} & 38.7\% & 29.3\% \\
			SAFA \cite{shi2019spatial} & - & 31.5\% \\
			SMD \cite{zhu2023simple} & - & 32.2\% \\
			\hline
		\end{tabular}
		\label{tab:model}
	}%
	\hfill%
	\parbox[t][][t]{.25\linewidth}{%
		\centering
		\caption{Comparison of different embedding dimensions.}
		\vspace{7.25mm} %
		\begin{tabular}[t]{l|r}
			\hline
			\rule{0pt}{2.2ex}Embedding & R@$1{<}50$m \\
			dimension & \\
			\hline
			\rule{0pt}{2.2ex}512 & 45.7\% \\
			1024 & 48.3\% \\
			2048 & 50.0\% \\
			4096 & 50.4\% \\
			\hline
		\end{tabular}
		\label{tab:embed-dim}
	}
\end{table}

We compare our MHA-based pooling with other pooling layers previously used in CVGL \cite{shi2019spatial,deuser2023sample4geo,zhu2023simple}. For fair comparison, we use the same encoder and embedding dimensions in all experiments. \Cref{tab:model} shows that our MHA-based pooling outperforms other layers both in the single LOD and multiple LOD setting. While Sample4Geo \cite{deuser2023sample4geo} reports state-of-the-art results on the VIGOR benchmark with a mean pooling layer, our experiments demonstrate that on our large-scale dataset more complex pooling layers provide a significant benefit.

The localization performance is further improved when increasing the embedding dimension (\cf \cref{tab:embed-dim}) and when decreasing the size of individual cells (\cf \cref{fig:searchregion}) at the cost of more memory requirement of the database and computational cost of the nearest neighbor search. We choose 1024 dimensions and cells with size $30\text{m} \times 30\text{m}$ as a trade-off between speed, memory and recall. The resulting reference database for the Massachusetts test split has a size of 104GB and allows querying a photo in less than 10ms using Faiss \cite{douze2024faiss}.

When using a Mercator-based cell layout as in VIGOR \cite{zhu2021vigor}, the model is trained on a range of cell sizes (\ie highlighted regions in \cref{fig:searchregion} that reflect the latitudes of selected states in the US and Germany). Models trained with our consistent cell layout achieve a comparable or higher recall than Mercator-based cells for test sets with any given cell size.

\section{Conclusion}

We present a method that is able to predict the geolocation of \mbox{street-view} images by matching against a database of aerial reference imagery. Our work for the first time allows localizing photos (1)~in large-scale search regions such as the entire state of Massachusetts, (2)~in challenging real-world scenarios with \mbox{consumer-grade} cameras and no additional information such as the camera parameters or compass angle, and (3)~without requiring data from the target region for training. We demonstrate the feasibility of our approach on a large and diverse crowd-sourced dataset and make the code publicly available.

\bibliographystyle{splncs04}
\bibliography{main}

\newpage

{\hspace{24mm}\Large\textbf{Supplementary Material}}

\section*{S1\quad Baselines}

We report the main results in our paper on the Massachusetts test split of our dataset. We reimplement SAFA \cite{shi2019spatial}, TransGeo \cite{zhu2022transgeo} and Sample4Geo \cite{deuser2023sample4geo} as baselines with the cell layout and choice of aerial images as in VIGOR \cite{zhu2021vigor}. The following list provides a detailed description of how these methods are adapted to our problem setup.
\begin{itemize}
	\item For SAFA and Sample4Geo, we use the same encoder architecture ConvNeXt-B \cite{liu2022convnet}. While TransGeo is originally introduced with a ViT-S encoder, we increase the size to ViT-B with 87M parameters resulting in a similar size as ConvNeXt-B with 89M parameters \cite{liu2022convnet}.
	\item For Sample4Geo, we use shared weights for the street-view and aerial encoders.
	
	\item We apply all models on a single aerial image at twice the size of the cell as in VIGOR.
	\item We use a resolution of $640 \times 480$ for the street-view images and $384 \times 384$ for the aerial images as in our work.
	\item We use a Mercator-based layout of the search region as in VIGOR such that cells in Massachusetts have a size of roughly $30\text{m} \times 30\text{m}$.
	
	\item Since the HEM strategies in Sample4Geo and TransGeo are designed for training on smaller datasets over multiple epochs, we use our semi-infinite mining strategy instead. For TransGeo, we reduce the cluster size to 2, and utilize 15 independent clusters per batch for a total batch size of $b = 30$. SAFA is implemented without HEM.
	
	\item Since Sample4Geo relies on a larger batch size, we increase it from 30 to 90 and train for 66k instead of 200k iterations. We
	use a learning rate of \mbox{$3 \times 10^{-4}$} according to the linear scaling rule \cite{goyal2017accurate}.
	\item We use a smaller learning rate of \mbox{$3 \times 10^{-5}$} in TransGeo which uses a ViT encoder.
	\item We use a symmetric cross-entropy loss for Sample4Geo, and a soft-margin triplet loss for SAFA and TransGeo.
	\item We use the same optimizer and learning rate schedule for the baselines as in our method. We do not employ Sharpness-Aware Minimization \cite{kwon2021asam,zhu2022transgeo} in our method or the baselines for fair comparison and since it doubles the computational cost of a training run.
	
	\item TransGeo proposes a refinement after the regular training where portions of the aerial input images that are deemed uninformative are removed, and the remaining portions are included with a higher resolution. This requires predicting and storing masks for all aerial input images in the dataset and training for additional epochs to fine-tune on the cropped images. Since the relative improvement of employing the refinement step is small (\eg 59.8\% to 61.5\% on the VIGOR same-area split) and incurs a much higher cost in our semi-infinite data setup than in a multi-epoch setup where images are seen more than once, we train TransGeo without this refinement.
\end{itemize}

\section*{S2\quad Dataset}

\begin{figure}[ht!]
	\subcaptionbox{North Rhine-Westphalia}
	{\includegraphics[height=3.6cm]{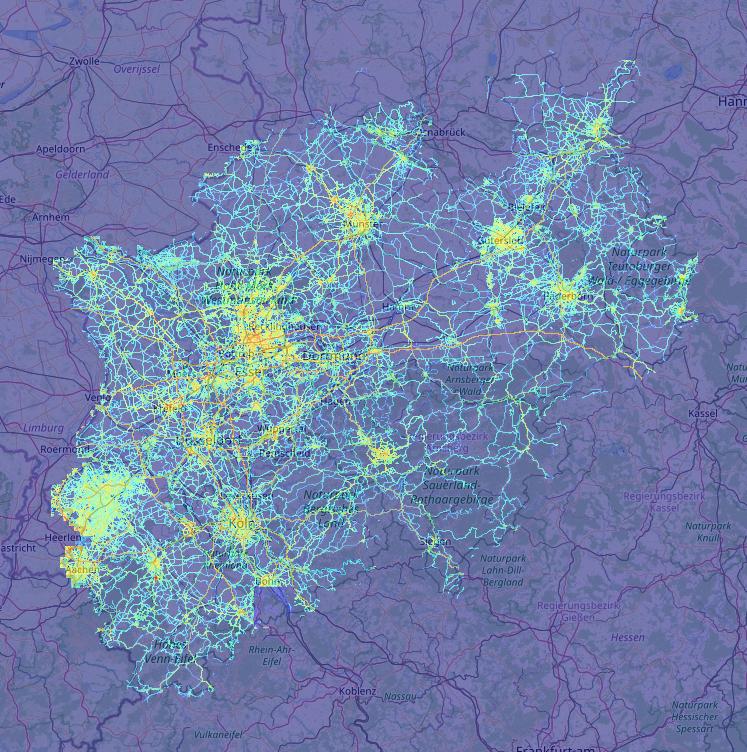}}
	\hfill
	\subcaptionbox{Berlin \& Brandenburg}
	{\includegraphics[height=3.6cm]{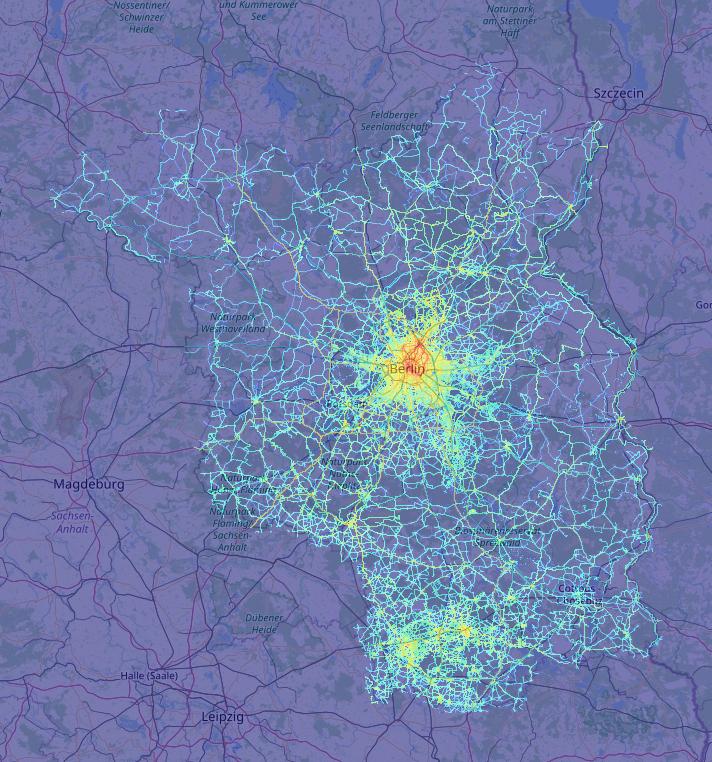}}
	\hfill
	\subcaptionbox{Saxony}
	{\includegraphics[height=3.6cm]{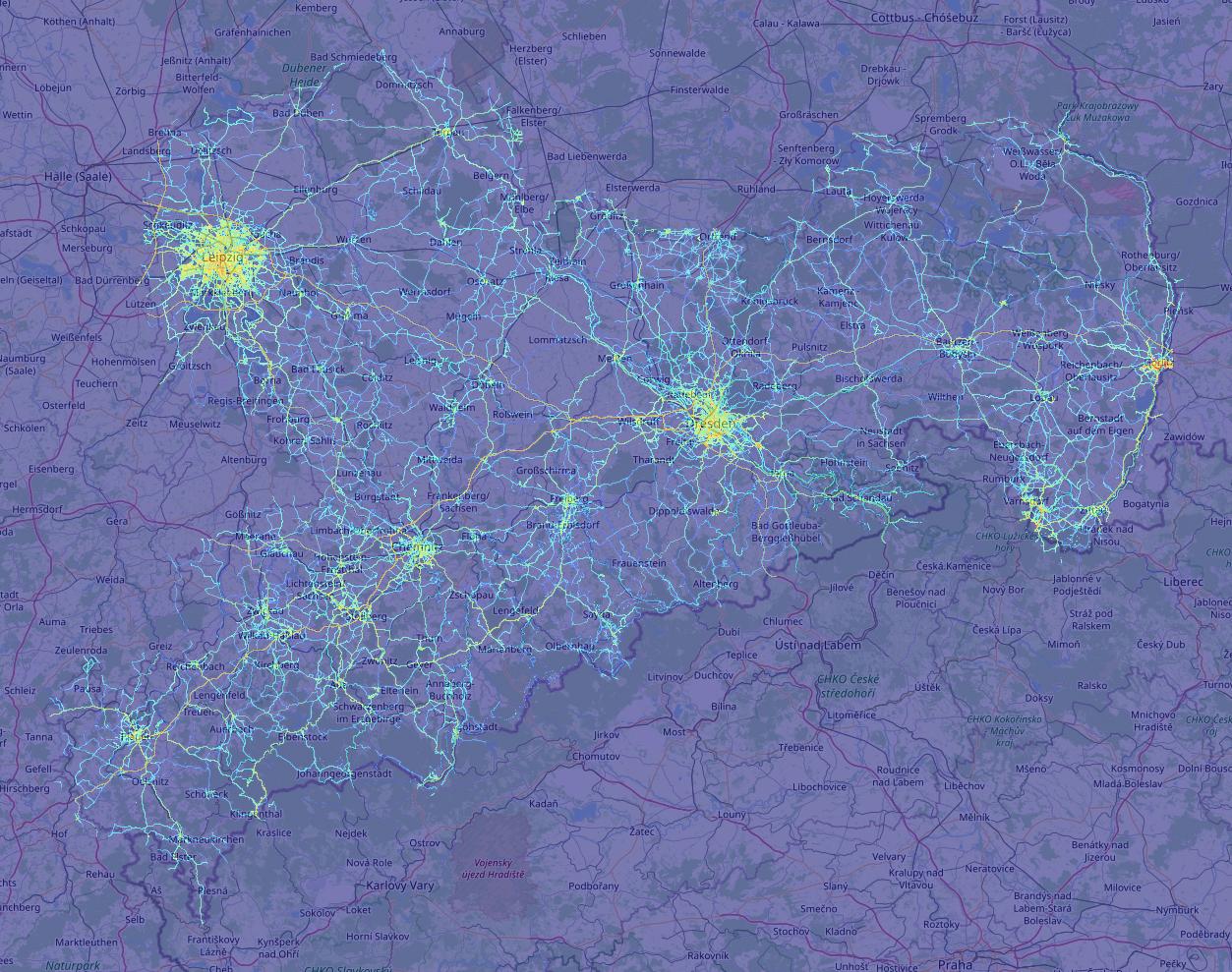}}\vspace{1mm}
	
	\subcaptionbox{Massachusetts}
	{\includegraphics[height=3.11cm]{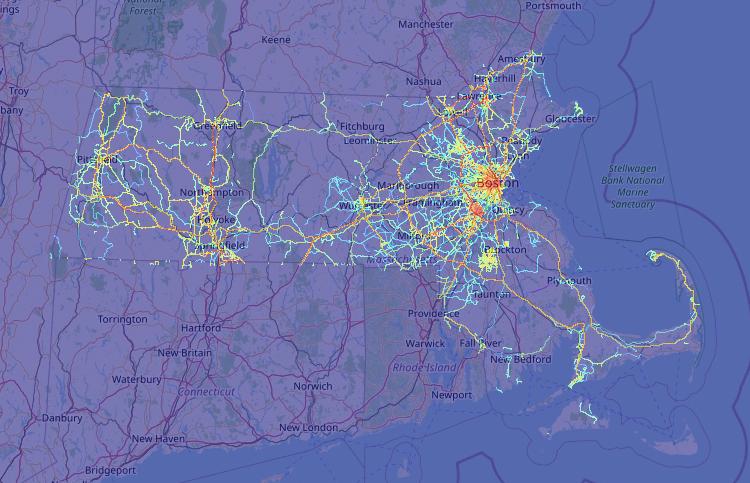}}
	\hfill
	\subcaptionbox{North Carolina}
	{\includegraphics[height=3.11cm]{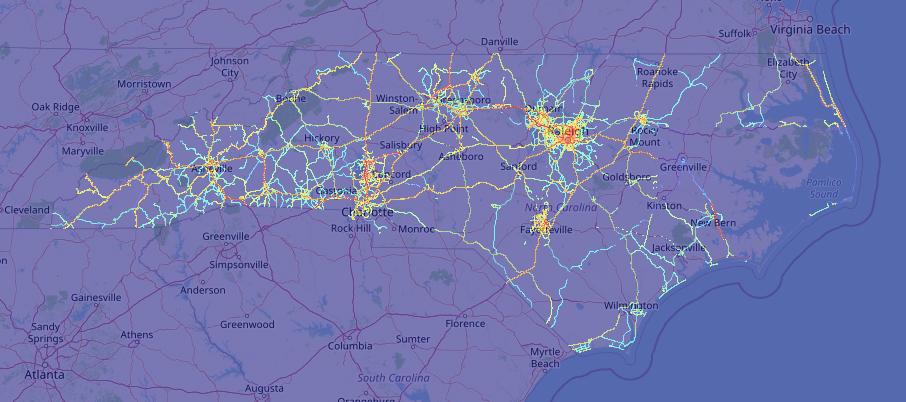}}\vspace{1mm}
	
	\subcaptionbox{Washington, D.C.}
	{\includegraphics[height=4.17cm]{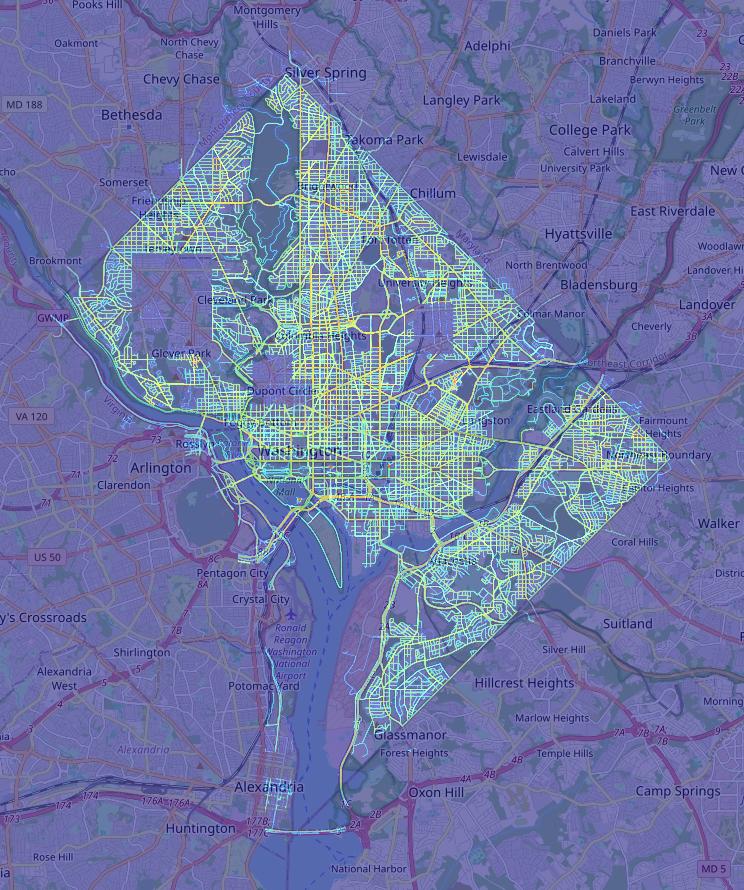}}
	\hfill
	\subcaptionbox{All states}
	{\includegraphics[height=4.17cm]{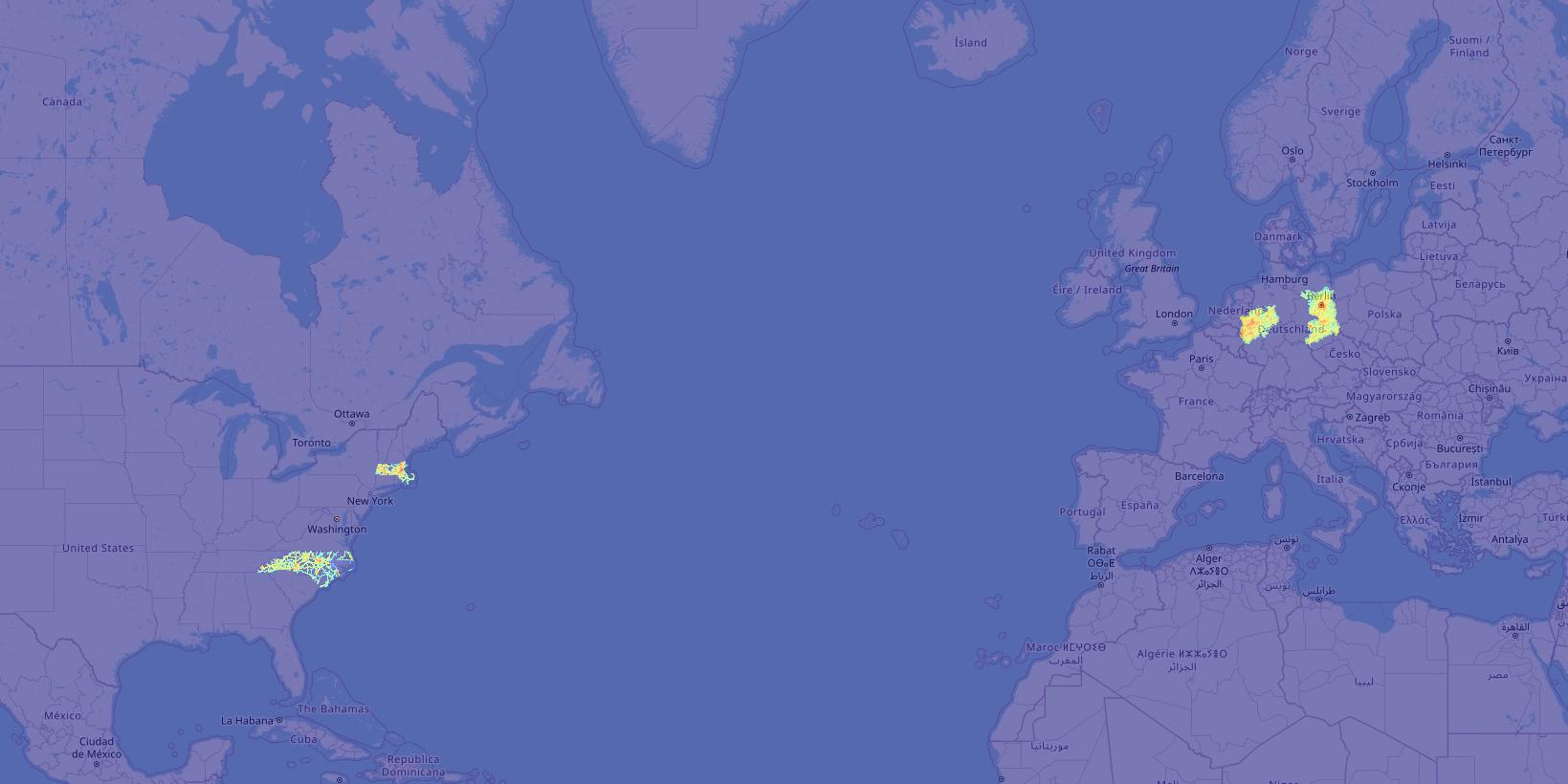}}
	
	\caption{Coverage of street-view photos in our dataset. The color indicates the number of photos per pixel (blue: zero photos, red: maximum number of photos in the subfigure). Map data: OpenStreetMap \cite{OpenStreetMap}.\label{fig:datasets}}
\end{figure}

All data used in this work are publicly available. The street-view data is downloaded via the Mapillary API for the selected regions in the US and Germany. We skip panoramic images and resize and pad all images to $640 \times 480$. We do not apply additional preprocessing steps. \cref{fig:datasets} provides a visualization of the geographical coverage of all street-view images used in this work. The aerial imagery is downloaded from the orthophoto providers of the respective states.

We provide download scripts and the IDs of all street-view images used in this work at the following URL:\\
\url{https://github.com/fferflo/statewide-visual-geolocalization}.

We annotate photos in all figures with the attribution: \textit{USER/IMAGE\_ID}. The image ID allows looking up the respective photo in the Mapillary service using the following URL:\\ \url{https://www.mapillary.com/app/?pKey=IMAGE_ID}.

\section*{S3\quad Search Region and Cell Sizes}

Our proposed search region layout consists of roughly square-shaped cells of size $l \times l$ over a spherical model of the earth with radius $r$. The approximation error is estimated as follows.
Given a longitudinal band of cells around the globe between the latitudes $\phi_a$ and $\phi_b = \phi_a + \frac{l}{r}$, the longitudinal circumference at the lower and upper end of the band is defined as:
\begin{equation}
	c_{\{a|b\}} = 2 \pi r \cos (\phi_{\{a|b\}})
\end{equation}
Each cell represents an isosceles trapezoid in a locally flat space, with a ratio between the shorter and longer side of $k=\frac{\min\{{c_a, c_b}\}}{\max\{{c_a, c_b}\}}$. Near the equator, the cells are roughly square shaped with $k \approx 1$, while near the poles they are roughly triangle shaped with $k \approx 0$. When using cells of size $l = 30\text{m}$ and latitudes with $-85.06^\circ < \phi < 85.06^\circ$ (\ie the maximum range depicted by common web maps \cite{epsg3857}), the deviation of the two opposing sides is bounded by $k > 1 - \epsilon$ with \mbox{$\epsilon \approx 6.3 \cdot 10^{-4}$}, \ie $1.9$cm for $l=30$m.

\section*{S4\quad Discussion}

\begin{figure}[t]
	\centering
	\includegraphics[width=\linewidth]{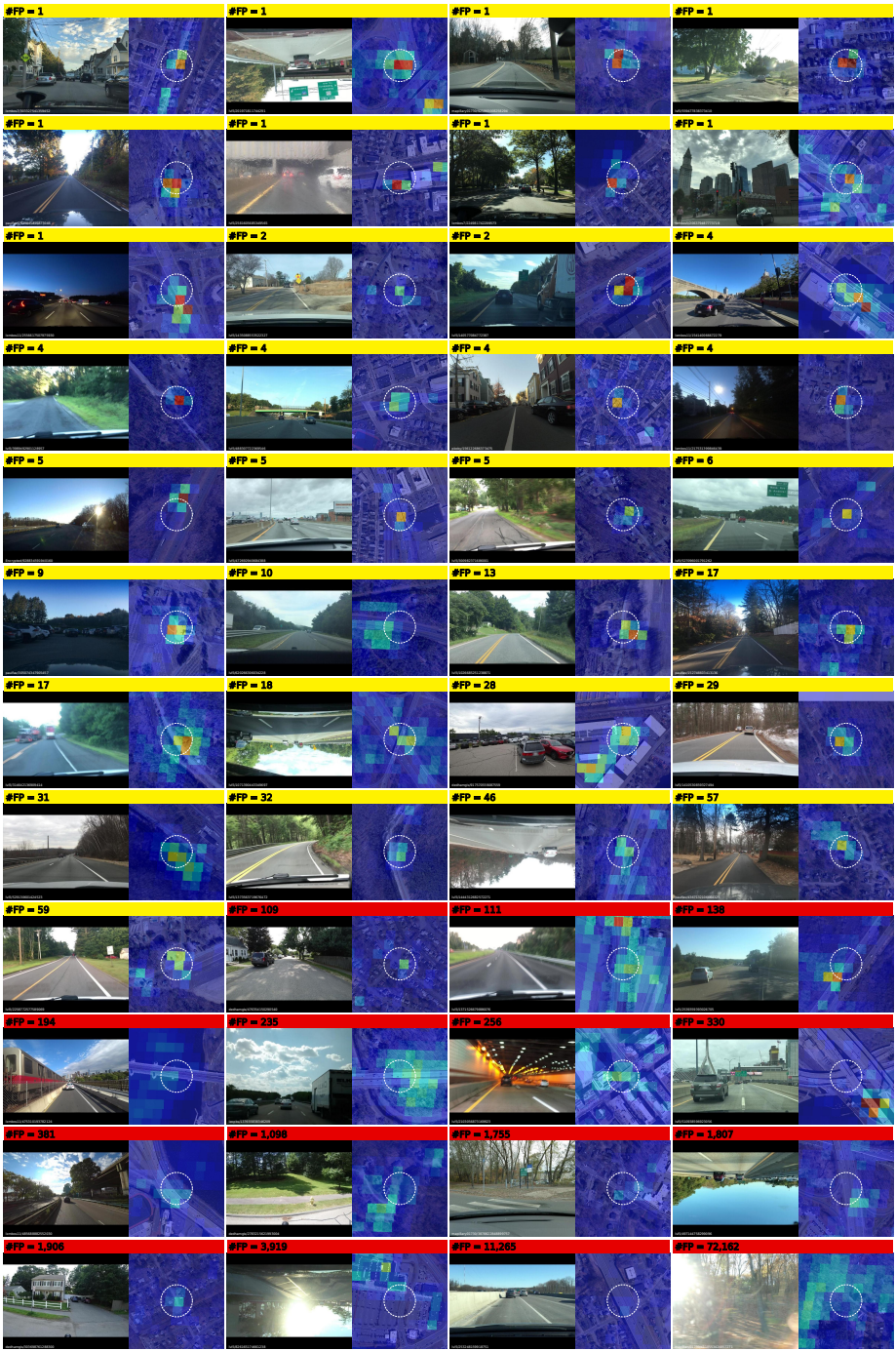}
	\caption{Randomly selected failure cases with at least one false positive (FP). The search region's color indicates the predicted score for possible camera locations. The white circle delinates the 50m radius around the ground-truth position.}
	\label{fig:failure}
\end{figure}

\paragraph{Limitations} \Cref{fig:failure} shows 48 randomly chosen failure cases in the Massachusetts test split containing at least one false positive cell predicted by the model. We identify two types of failures:
\begin{enumerate}
	\item Images that cannot be evaluated correctly, \eg due to image artifacts, errors in the ground-truth location of more than 50m, or missing overlap between \mbox{street-view} photo and aerial image, \eg due to being captured in tunnels or indoors.
	\item Images that potentially allow for a correct prediction, but are incorrectly localized by the model, \eg due to other locations in the dataset that are easy to confuse with the true location of the query. We also find that the model's performance is reduced on images with a large deviation from the frontal street-view perspective that is heavily overrepresented in the \mbox{crowd-sourced} data on Mapillary.
\end{enumerate}

\paragraph{Application of our method to existing datasets} Most datasets in the field of CVGL \cite{workman2015wide,liu2019lending,vo2016localizing,zhang2023cross,vyas2022gama} do not represent a many-to-one matching problem with a dense coverage of a test region that our method could be applied to. VIGOR \cite{zhu2021vigor} is the only existing dataset that provides such coverage, but contains no interface to retrieve aerial image patches other than those included in their predefined layout. Applying our method to this dataset would require stitching the images back together to form a large contiguous aerial image of the entire region, and then resampling and extracting patches following our layout with a consistent resolution and at multiple LOD. However, this would still leave many cells near the region boundary with large parts of the zoomed out LODs missing, since our aerial images cover a much larger surrounding area per cell than VIGOR. New aerial imagery for these regions is also not publicly available for free and under permissive licenses - unlike for our test regions which we chose for this reason.

\paragraph{Potential negative societal impact} Visual geolocalization raises potential privacy concerns, \eg when localizing images shared on social media. Our method is trained on and limited to street-view images captured in public outdoor spaces and does not allow localizing arbitrary personal photos. However, the authors believe that awareness of the potential dangers of sharing personal data online is crucial.

\end{document}